\documentclass[runningheads]{llncs}
\usepackage[T1]{fontenc}
\usepackage{graphicx}

\usepackage{booktabs}
\usepackage{marvosym}
\usepackage{hyperref}
\usepackage{overpic}
\usepackage{bm}
\usepackage{amsmath,array,arydshln}
\usepackage{multicol}
\usepackage{multirow}
\usepackage{hyperref}
\usepackage{amsfonts}
\usepackage{cite}
\usepackage{ifsym}
\usepackage{subfigure}

\begin{document}
\title{Hierarchical Vision-Language Learning for
Medical Out-of-Distribution Detection}
%
%
\author{Runhe Lai\inst{1,2,5} \and
Xinhua Lu\inst{1,2,5} \and
Kanghao Chen\inst{3}\textsuperscript{(\Letter)}\and
Qichao Chen\inst{4}\and
Wei-Shi Zheng \inst{1,2,5}\and
Ruixuan Wang\inst{1,2,5}\textsuperscript{(\Letter)} }

\authorrunning{R. Lai et al.}
%
\institute{School of Computer Science and Engineering, Sun Yat-sen Univerisity, Guangzhou,China\and
Peng Cheng Laboratory, Shenzhen, China\and
Hong Kong University of Science and Technology (Guangzhou), Guangzhou, China \and
University of Nottingham Malaysia, Semenyih, Malaysia\and
Key Laboratory of Machine Intelligence and Advanced Computing, MOE, Guangzhou, China \\
\email{kchen879@connect.hkust-gz.edu.cn, wangruix5@mail.sysu.edu.cn}
}
%
\maketitle              
\begin{abstract}
\begin{sloppypar}
In trustworthy medical diagnosis systems, integrating out-of-distribution (OOD) detection aims to identify unknown diseases in samples, thereby mitigating the risk of misdiagnosis. 
In this study, we propose a novel OOD detection framework based on vision-language models (VLMs), which integrates hierarchical visual information to cope with challenging unknown diseases that resemble known diseases.
Specifically, a cross-scale visual fusion strategy is proposed to couple visual embeddings from multiple scales.
This enriches the detailed representation of medical images and thus improves the discrimination of unknown diseases.
Moreover, a cross-scale hard pseudo-OOD sample generation strategy is proposed to benefit OOD detection maximally.
Experimental evaluations on three public medical datasets support that the proposed framework achieves superior OOD detection performance compared to existing methods.
The source code is available at https://openi.pcl.ac.cn/OpenMedIA/HVL.
\end{sloppypar}
\end{abstract}

\section{Introduction}

An intelligent medical diagnosis system is typically designed to recognize some disease categories after the system are trained on known in-distribution (ID) data. 
However, in real-world scenarios, the system would often encounter medical images of unknown diseases which are called out-of-distribution (OOD) data.
OOD detection aims to identify unknown diseases and reject uncertain predictions, thereby improving the system's robustness.

Multiple developed methods~\cite{ViM,NPOS,Evidence,GEN,longtailOOD,zhou2022rethinking} have demonstrated impressive OOD detection performance in the medical image and natural image domains.
However, these methods are uni-modal 
and often incapable of detecting challenging OOD data that are visually similar to ID classes. 
In contrast, very recent studies~\cite{SCT,LoCoOp,GalLop,eok} start to explore pre-trained vision-language models (VLMs) like CLIP~\cite{CLIP} for OOD detection, which use text guidance to learn a better decision boundary between ID and OOD data.
In particular, local image regions such as background regions in ID images are commonly adopted to construct pseudo-OOD samples~\cite{eok,LoCoOp,SCT}. However, in medical images, background regions are often clearly different from lesion regions, while lesions of unknown diseases in OOD images could be very similar to some learned known (ID) diseases. This makes existing local region-based methods ineffective for medical image analysis.

In this paper, we propose a novel hierarchical vision-language learning framework for OOD detection tailored for medical image analysis. 
Unlike previous studies~\cite{eok,GalLop} that analyze local regions and the original image as two separate parts of the information source, 
this study proposes a cross-scale visual fusion to couple local details with global information from the original image for a comprehensive understanding of medical images.
Moreover, this study proposes a cross-scale hard pseudo-OOD sample generation strategy for OOD detection.
Specifically, a novel entropy gain is applied to construct hard pseudo-OOD visual embeddings around boundary areas of lesions in ID images. 
By learning to identify hard pseudo-OOD visual embeddings from the patch level to the whole image level, the VLM can achieve better OOD detection performance. 
Extensive experiments on three datasets with diverse sample distributions support our framework achieves {state-of-the-art} OOD detection performance, thus improving reliability for clinical applications.
Our contributions are summarized below.
\begin{itemize}
    \item A novel hierarchical vision-language learning framework based on VLMs for medical OOD detection;
    \item A cross-scale visual fusion method for coupling both local and global information to promote the detection of unknown diseases;
    \item A hard pseudo-OOD generation strategy across scales for OOD detection.
\end{itemize}

\noindent\textbf{Relation to concurrent work: }A concurrent work GLAli~\cite{GLAli} is proposed to unify ID classification and OOD detection in few-shot scenario by incorporating visually guided text refinement, local contrastive learning, and multi-scale image–text alignment, enhancing performance across both tasks. 
Unlike GLAli, our method focuses on modeling interactions among hierarchical visual embeddings under pseudo-OOD guidance to better tackle the near-OOD challenge.

\section{Method}
This study aims to improve OOD detection ability of a diagnostic system which is trained only with available ID data. During inference, the model is expected to diagnose all known diseases and detect unknown diseases.

\begin{figure}[t]
\begin{center}
\includegraphics[width=0.9\textwidth]{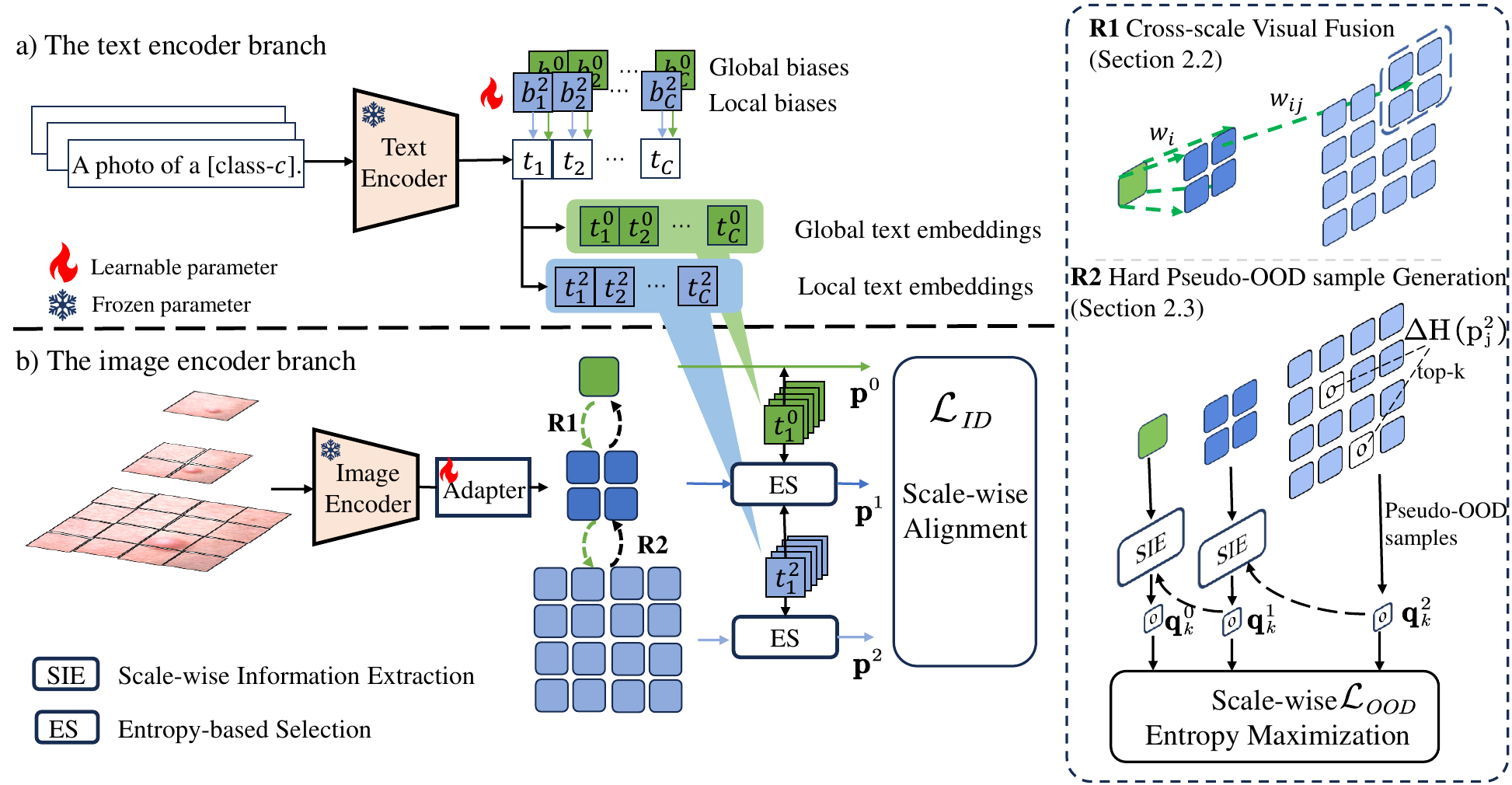}
\end{center}
   \caption{Overview of our proposed HVL framework.
\textbf{Left}:  Visual embeddings from the original image and its patches are fused across scales and different learnable biases are added to text embeddings to align with visual embeddings.
\textbf{Right}: 
(R1). Lower-scale visual embeddings progressively enhance higher-scale visual embeddings. 
(R2). High-scale hard pseudo-OOD embeddings are selected via entropy gain and propagated to lower scales.
Then, these pseudo-OOD embeddings are used for model training.  }
\label{figure_framework}
\end{figure}

\subsection{Framework Overview}

Our proposed framework called HVL is illustrated in Figure~\ref{figure_framework}. 
In the image encoder branch, we construct hierarchical inputs by upsampling the image and partitioning it into multi-scale patches. 
At the output side of the image encoder followed by the learnable adapter, there is a visual fusion route (R1) and a hard pseudo-OOD embedding generation route (R2) with different delivery directions. 
For the visual fusion route R1, inspired by the combination of local pathology and global information for diagnostic decision-making in common diagnostic scenarios,
a cross-scale visual fusion is proposed to couple visual embeddings from multiple scales in a coarse-to-fine manner as shown in Figure~\ref{figure_framework}, R1. 
The enhanced embeddings support the model to explore the visual information from a contextual perspective rather than a simple combination of local region information and global information as in previous work~\cite{eok,GalLop}.
In the pseudo-OOD sample generation route (Figure~\ref{figure_framework}, R2), 
we propose a strategy to select hard pseudo-OOD regions from boundary areas surrounding lesions, and then the model learns to differentiate these hard-pseudo OOD regions from ID lesions at different scales, enabling it to identify unknown diseases through multiple scales. 
For the text encoder branch (Figure~\ref{figure_framework}, a), we propose adjusting text embeddings with different learnable biases to align with hierarchical visual embeddings.

\subsection{Cross-Scale Visual Fusion and Hierarchical Feature Alignment}\label{sec:id}

Previous studies~\cite{LoCoOp,LSN,SCT,negprompt,id-like} employ CLIP’s 
image encoder, where the entire image is encoded as an image-level presentation for OOD detection. 
Thus, these methods fail to extract fine-grained details 
of local regions 
and may not accurately detect unknown diseases if the corresponding lesion regions are small in images. 
To address this limitation, we propose partitioning each image into multi-scale patches to extract more discriminative detail information. 
Specifically, as shown in Figure~\ref{figure_framework},   firstly $n \times$ and $2n \times$ upsampling (e.g., $n=2$) are respectively applied to the original image $\mathbf{x}$ to obtain the mid-scale and high-scale inputs, and then the two upsampled images are partitioned into $n\times n$ and $2n\times 2n$ patches respectively. 
Visual embeddings of the original image and the partitioned patches 
are extracted using the CLIP's image encoder $f_I$ followed by a learnable adapter.
Formally, the original image $\mathbf{x}$ is encoded in $\mathbf{v} = f_I(\mathbf{x})\in \mathbb{R}^{d \times 1}$, and then adjusted by the adapter to the visual embedding
$\mathbf{u}^0 = {\rm ReLU}(\mathbf{v}^{\rm T} \mathbf{W}) + \mathbf{v}$,
where $\mathbf{W}\in\mathbb{R}^{d\times d}$. Similarly,
patches from the mid- and high-scale inputs are encoded into visual embeddings 
$\{\mathbf{u}^1_i\}^{n\times n}_{i=1}$ and $\{\mathbf{u}^2_j \}_{j=1}^{2n\times 2n}$.
These initial hierarchical embeddings of the image $\mathbf{x}$ and will be improved as follows.

In disease diagnosis scenarios, it is common for human doctors to combine local 
pathology with global information to make a diagnosis. Intuitively, coupling 
global and local information can help 
understand the context of local details in the original image, which promotes the 
reliable detection of unknown diseases even if they are visually similar to certain known diseases. 
Inspired by such observation, we propose a cross-scale visual feature fusion strategy to complement global visual information for OOD detection of unknown diseases.  
Specifically, the embedding of each higher-scale patch is augmented with the corresponding lower-scale patch's embedding in a coarse-to-fine manner, i.e.,
\begin{equation}
    \hat{\mathbf{u}}_{i}^1 = \mathbf{u}^1_i + \cos(\mathbf{u}_i^1, \mathbf{u}^0)\cdot \mathbf{u}^0,
    \quad
    \hat{\mathbf{u}}_{j}^2 = \mathbf {u}_{j}^2 + \cos(\mathbf{u}_{j}^2,\hat{\mathbf{u}}_{j^*}^1)\cdot \hat{\mathbf{u}}_{j^*}^1,
\end{equation}
where $\cos(\cdot,\cdot)$ is the cosine similarity function, and $\hat{\mathbf{u}}_{j^*}^1$ represents the augmented embedding of the mid-scale patch whose further partitions include the higher-scale patch associated with the embedding $\mathbf {u}_{j}^2$. 
The weighting strategy based on cosine similarity ensures that the lower-scale information as a context will be passed to embeddings of higher-scale patches, while the information in higher-scale patch embeddings which is more different from the lower-scale information will be well preserved in the augmented higher-scale patch embeddings.  

These refined higher-scale visual embeddings $\hat{\mathbf{u}}_{i}^1$'s and $\hat{\mathbf{u}}_{j}^2$'s, together with the global embedding $\mathbf{u}^0$, will be used in model training and inference. However, some mid- and high-scale patches may only contain normal tissues, and such disease-relevant in general should not be used to recognize the disease category of the original 
image. Therefore, it is necessary to find and then remove those disease-irrelevant patches when classifying each image. 
In our proposed framework HVL, the text encoder branch is used not only to help classify each image but also to help find disease-irrelevant patches as detailed below.

In the text encoder branch, let $\mathbf{z}_c$ denote the text description ``a photo of a [\textit{class-c}]'', 
where \textit{class-c} represents the class name or description for class $c$.  
The text embedding $\mathbf{t}_c$ 
for class $c$ from the CLIP's text encoder can be obtained as $\mathbf t_c=f_T(\mathbf{z}_c)\in \mathbb{R}^{d \times 1}$. Denote by $\mathbf{t} = [\mathbf{t}_1,...,\mathbf{t}_C]\in \mathbb{R}^{d\times C}$ the text embedding collection for all $C$ ID classes. 
Considering that visual embeddings 
contain visual information of different scales, it is probably unsuitable to match multi-scale visual embeddings with 
the initial text embedding. Therefore, we propose adjusting the text embeddings with learnable biases. 
Specifically, learnable biases $\mathbf{b}^0, \mathbf{b}^2\in \mathbb{R}^{d\times C}$ are each added to $\mathbf t$ to obtain 
$\mathbf{t}^0 = \mathbf{t} + \mathbf{b}^0$ and $\mathbf{t}^2 = \mathbf{t} + \mathbf{b}^2$ which will be used to match visual embeddings from the original input and the high-scale patches, respectively. 
The mid-scale text embeddings are obtained as $\mathbf{t}^1 = \mathbf{t} + \mathbf{b}^1 = [\mathbf{t}^1_1,...,\mathbf{t}^1_C]$, where the bias $\mathbf{b}^1=\frac{1}{2}(\mathbf{b}^0+\mathbf{b}^2)$ is not independently learnable but from the 
biases $\mathbf{b}^0$  and $\mathbf{b}^2$ 
for coupling information across scales. 
Then, scale-wise alignment between visual embeddings and text embeddings can be 
computed. For example, the degree of alignment between the mid-scale visual embedding $\hat{\mathbf{u}}^1_i$ and the text embedding $\mathbf{t}^1_c$ of class $c$ is $\cos(\hat{\mathbf{u}}^1_i, \mathbf t_c^1)$. 
Such alignment can be used to estimate the probability of classifying 
$\hat{\mathbf{u}}^1_i$ as class $c$, i.e., 
\begin{equation}
    p( y= c|\hat{\mathbf{u}}^1_i) =\frac{{\rm exp}({\cos}(\hat{\mathbf{u}}^1_i, \mathbf t_c^1) / \tau)}{\sum_{c'=1}^C{\rm exp}({\cos}(\hat{\mathbf{u}}^1_i, \mathbf t_{c'}^1)/ \tau)} \,,
    \label{eq:alignment}
\end{equation}
where $\tau$  is a fixed temperature scaling hyper-parameter. Since disease-irrelevant normal tissues often appear in images of all disease classes, the probability of classifying one disease-irrelevant patch into each class will be more or less similar to each other. With this consideration, the entropy of the probability distribution over all $C$ classes can be used to determine which patches are likely disease-irrelevant. Specifically, let $\mathbf{p}^1_i \in \mathbb{R}^{C \times 1}$ denote the probability distribution 
for the mid-scale patch associated with $\hat{\mathbf{u}}^1_i$, where each component in $\mathbf{p}^1_i$ is computed based on Equation~\ref{eq:alignment}.
The average entropy over all mid-scale patches can be obtained as $\overline{\rm H}^1 = \sum_i{\rm H}(\mathbf{p}_i^1)$, with ${\rm H}(\mathbf{p}_i^1)$ representing the 
entropy of 
$\mathbf{p}^1_i$. 
Since disease-irrelevant patches often have 
higher entropy, those patches whose 
entropy is higher than 
$\overline{\rm H}^1$ will be estimated as disease-irrelevant patches and therefore not considered for 
prediction. Consequently, the prediction probability distribution 
based the mid-scale patches can be estimated by the average ${\mathbf p}^1=\frac{1}{n \times n}\sum_{i=1}^{n \times n} \mathbb{I}({\rm H}(\mathbf{p}_i^1) \leq \overline{\rm H}^1 ) \cdot \mathbf{p}_{i}^1$, where $\mathbb{I}(\cdot)$ represents the indicator function. Similarly, the prediction probability distribution $\mathbf{p}^2$ based on the high-scale patches can be obtained. Overall, the model can be trained by minimizing  three cross-entropy loss terms $\mathcal{L}_{CE}$ based on scale-wise image-text alignment, i.e.,
\begin{equation}\label{eq:ID}
    \mathcal{L}_{ID}=\mathbb{E}_{(\mathbf{x}, y)\sim \mathcal{D}^{in}} \left\{\mathcal{L}_{CE}(\mathbf p^0,y)+\mathcal{L}_{CE}({\mathbf p}^{1},y)+\mathcal{L}_{CE}({\mathbf p}^2,y)  \right\} \,,
\end{equation}
where $\mathcal{D}^{in}$ represents the training set of all $C$ ID classes and $\mathbb{E}$ for expectation. $\mathbf{ p}^0$ is obtained based on the global visual embedding $\mathbf{u}^0$ and text embeddings $\mathbf{t}^0$.

\subsection{Hard Pseudo-OOD Sample Generation} \label{section-OOD} 
To improve OOD detection performance, previous efforts leverage ID-irrelevant local regions such as background in images as pseudo-OOD data for model training~\cite{LoCoOp,SCT,eok}. 
However, in medical images, normal tissues in background regions are often quite distinct from lesion regions, while the lesions of unknown diseases in 
OOD images are often similar to ID disease lesions. Therefore, using pure background regions as pseudo-OOD data may not be effective in improving OOD detection in medical scenarios. 

Generally, rather than the disease-irrelevant 
regions, boundary areas surrounding lesions are more similar to lesion regions because of their tissue-related information.
To use the regions from lesion boundary areas as hard OOD samples for model training, a novel entropy gain based strategy is proposed to locate these regions. 
The basic idea is to find those mid-scale patches that likely contain lesion information, while at least one high-scale patch from the partitions of each selected mid-scale patch are less likely to contain lesions. 
Such high-scale patches from boundary areas surrounding lesions can be selected based on the entropy of mid-scale regions and the corresponding high-scale patches.
Specifically, the selected high-scale patches have higher entropy gain ${\rm \Delta H}(\mathbf p_{j}^2) = {\rm H}(\mathbf p_{j}^2)-{\rm H}(\mathbf p_{j^*}^1)$,
where the entropy term $-{\rm H}(\mathbf p_{j^*}^1)$ encourages selecting the high-scale patches from the partitions of the regions near the lesions, while entropy ${\rm H}(\mathbf p_{j}^2)$ help avoid selecting the high-scale patches containing lesions. 
Suppose top-$K$ visual embeddings $\{{\rm \textbf{q}}^2_k\}^K_{k=1}\subset \{\hat{\mathbf{u}}^2_j\}^{2n\times 2n}_{j=1}$ of high-scale patches are selected. These top-$K$ visual embeddings of high-scale patches will be used as hard pseudo-OOD embeddings at the high-scale level to help improve the model's OOD detection performance. To construct hard pseudo-OOD visual embeddings which would help increase the ID-OOD discrimination ability of text embeddings at lower-scale levels, 
each selected $\mathbf{q}^2_k$ is progressively fused with all embeddings from each scale weighted by cosine similarity as follows,
\begin{equation}\label{eq:refinement}
    \mathbf q_k^1=\mathbf q_{k}^2+\frac{1}{n\times n}\sum_{i=1}^{n\times n}{\cos}(\mathbf q_{k}^2,\hat{\mathbf{u}}^1_i)\cdot \hat{\mathbf{u}}^1_i,\quad
    \mathbf q_{k}^0=\mathbf q_{k}^1+{\cos}(\mathbf q_{k}^1,\mathbf {u}^0)\cdot \mathbf {u}^0 \,.
\end{equation}
In this way, some lesion information at each lower scale will be inserted into the visual embeddings $\mathbf q_{k}^0$ and $\mathbf q^1_k$, making the embeddings difficult to differentiate from the visual embeddings of real lesions, while the cosine similarity controls the proportion of lesion information in the hard embeddings $\mathbf q_{k}^0$ and $\mathbf q^1_k$ to a low level. With these hard pseudo-OOD visual embeddings at three scales, the model's OOD detection performance can be improved by the well-known outlier exposure strategy~\cite{OE}, i.e., maximizing entropy (or minimizing negative entropy) of prediction probability distribution for pseudo-OOD data with the loss $\mathcal{L}_{OOD}$
\begin{equation}\label{eq:ood}
    \mathcal{L}_{OOD}=-\mathbb{E}_{\mathbf x\sim \mathcal{D}^{in}}\{\frac{1}{K}\sum_{k=1}^{K}({\rm H}( \mathbf p_{k}^0)+{\rm H}(\mathbf p^{1}_k)+{\rm H}(\mathbf p^{2}_k))\} \,,
\end{equation}
where $\mathbf p^{1}_k$ represents the prediction probability distribution whose components are from the alignment between the visual embedding $\mathbf q^1_k$ and the text embeddings $\mathbf{t}^1$ (refer to Equation~\ref{eq:alignment}), and similarly for $\mathbf p^{0}_k$ and $\mathbf p^{2}_k$. Overall, the model will be trained by minimizing the combined loss $\mathcal{L}=\mathcal{L}_{ID}+\mathcal{L}_{OOD}$.

During model inference, for any test image, prediction distributions $\mathbf p^0,{\mathbf p}^1,{\mathbf p}^2$ 
from the three scales are averaged as the final prediction distribution $\mathbf p_{ID}=\frac{1}{3}(\mathbf p^0 + \mathbf p^1 + \mathbf p^2)$, 
where $\mathbf p^0,{\mathbf p}^1,{\mathbf p}^2$ are obtained as described in Section~\ref{sec:id}.
For OOD detection, the MSP~\cite{MSP} score, i.e., the maximum output component in the the final prediction distribution $\mathbf p_{ID}$, is used to identify OOD samples, 
with lower score suggesting that the test data is more likely an OOD sample.

\section{Experiment}
\subsection{Experimental Setup}
\textbf{Datasets:} We conducted experiments on the SD-198~\cite{SD198}, ISIC 2019~\cite{ISIC} for dermatology and NCT-CRC~\cite{NCT} datasets for histology respectively. 
For the SD-198 dataset, we used its subset Skin-40~\cite{Skin40} which contains 40 categories as ID classes and the collection of the remaining 158 categories as OOD data.
For the ISIC 2019 dataset, we selected NV, MEL, DF and VASC as ID categories with a long tail data distribution and called it ISIC-4, while the remaining four categories serve as OOD data.
Following the setup in previous work~\cite{Arewe}, we divided the NCT-CRC dataset into three OOD detection tasks, each of which designates three categories as OOD data, while the remaining categories serve as ID data. We report the average performance of these three tasks.
Note that we employed five-fold cross-validation on ISIC-4 benchmark while training all models on the official training set of Skin40/NCT-CRC with three different seeds.

\noindent \textbf{Implementation details:} The pre-trained CLIP (ViT-B/16~\cite{vit}) was used as backbone in all experiments.
The number of selected regions $K$ was set to 4 and the number of partitions $n$ was set to 2 by default. 
The model was trained up to 100 epochs using the Adam optimizer~\cite{Adam} with a learning rate of 0.002 and a mini-batch size of 32, following a cosine annealing schedule.

\begin{table}[tbh]
\caption{Performance comparison on the three benchmarks. 
VLM-based methods are marked with ``\checkmark'' while uni-modal methods are marked with `` - ''.
All values are percentages. 
The range of standard deviation is [0.13, 1.72].
}\label{tab1}
\resizebox{\linewidth}{!}{
\begin{tabular}{ l c ccc ccc ccc ccc}

    \toprule
 \multicolumn{2}{c}{\textbf{ID}}   &\multicolumn{3}{c}{\textbf{NCT-CRC}} &
\multicolumn{3}{c}{\textbf{ISIC-4}} & 
\multicolumn{3}{c}{\textbf{Skin40}} & 
\multicolumn{3}{c}{\textbf{Average}}\\
\cmidrule(lr){1-2} \cmidrule(lr){3-5} 
\cmidrule(lr){6-8} \cmidrule(lr){9-11} 
\cmidrule(lr){12-14}
\multicolumn{2}{c}{Method}    &
Acc & F$\downarrow$ & A$\uparrow$ & 
Acc & F$\downarrow$ & A$\uparrow$ & 
Acc & F$\downarrow$ & A$\uparrow$ & 
Acc & F$\downarrow$ & A$\uparrow$\\
\midrule
CoPR   & - 
& \underline{97.94}     & \underline{50.65}{\tiny $\pm$ 0.74}     &  88.77{\tiny $\pm$0.36}  
& \textbf{90.32}        &  68.78{\tiny $\pm$0.51}                 &  81.89{\tiny $\pm$0.93}
&  68.08                &  85.33{\tiny $\pm$0.89}                 &  69.72{\tiny $\pm$0.38}  
&  85.45                 &  68.25{\tiny $\pm$0.45}               &  80.13{\tiny $\pm$0.42}\\

ViM   & - 
& \underline{97.94}         &  54.33{\tiny $\pm$1.03}                &  88.24{\tiny $\pm$1.25} 
& \textbf{90.32}            &  65.13{\tiny $\pm$0.55}                & \underline{84.22}{\tiny $\pm$0.13}     
&  68.08                    & \underline{77.24}{\tiny $\pm$0.71}     &  70.67 {\tiny $\pm$0.35}
&  85.45                    & \underline{65.56}{\tiny $\pm$0.56}     & \underline{81.04}{\tiny $\pm$0.42} \\

GEN   & - 
&  \underline{97.94}  & 51.26{\tiny $\pm$0.92}               &  \underline{89.02}{\tiny $\pm$0.74}  
& \textbf{90.32}      &  \underline{63.71}{\tiny $\pm$0.34}  & 82.18{\tiny $\pm$0.62}  
&  68.08              &  86.38{\tiny $\pm$1.00}              &  63.65{\tiny $\pm$0.73} 
&  85.45              &  67.11{\tiny $\pm$0.87}              &  78.28{\tiny $\pm$0.72}\\

NPOS  & - 
&  95.25 &  57.21{\tiny $\pm$1.64}  &  86.02{\tiny $\pm$1.02} 
&  85.03 &  65.45{\tiny $\pm$1.03}  &  80.94{\tiny $\pm$1.18} 
&  71.32 &  81.70{\tiny $\pm$1.25}  &  \underline{71.85}{\tiny $\pm$0.83} 
&  83.87 &  68.12{\tiny $\pm$1.15}  & 79.60{\tiny $\pm$0.93}\\
\midrule

LoCoOp  & \checkmark    & 
97.68       &   54.38{\tiny $\pm$0.76}     & 85.61{\tiny $\pm$0.53}       & 
82.40       &   69.04{\tiny $\pm$0.83}     & 83.21{\tiny $\pm$0.94}       & 
63.30       &   85.03{\tiny $\pm$0.73}     & 68.03{\tiny $\pm$0.68}       & 
81.13       &   69.48{\tiny $\pm$0.71}     & 78.95{\tiny $\pm$0.66}     \\

SCT     & \checkmark & 
95.50   &   58.08{\tiny $\pm$0.51}     &     83.81{\tiny $\pm$0.25}       & 
83.10   &   67.90{\tiny $\pm$0.91}     &     81.21{\tiny $\pm$0.75}       & 
64.33   &   89.74{\tiny $\pm$0.59}     &     63.12{\tiny $\pm$0.42}       & 
80.97   &   71.90{\tiny $\pm$0.63}     &     76.05{\tiny $\pm$0.51} \\

LSN     &  \checkmark & 
96.97   &   62.30{\tiny $\pm$1.72}     &  83.87{\tiny $\pm$0.56}     &  
80.85   &   67.44{\tiny $\pm$1.66}     &  80.03{\tiny $\pm$1.47}     & 
65.15   &   84.05{\tiny $\pm$1.45}     &  67.18{\tiny $\pm$1.07}     & 
80.99   &   71.26{\tiny $\pm$1.59}     &  77.03{\tiny $\pm$1.22}\\

GalLop  &  \checkmark & 
98.11           &  59.91{\tiny $\pm$0.44}        & 87.27{\tiny $\pm$0.65}     & 
85.15           &  72.95{\tiny $\pm$0.95}        & 78.35{\tiny $\pm$0.77}     & 
\textbf{76.75}  &  86.83{\tiny $\pm$0.21}        & 69.08{\tiny $\pm$0.13}     & 
\textbf{86.67}  &  73.23{\tiny $\pm$0.72}        & 78.23{\tiny $\pm$0.53}\\

Ours    & \checkmark & 
\textbf{98.30}      & \textbf{37.59}{\tiny $\pm$0.82}    & \textbf{90.95}{\tiny $\pm$0.65}    & 
\underline{88.43}   & \textbf{56.12}{\tiny $\pm$0.73}    & \textbf{85.75}{\tiny $\pm$0.56}    & 
\underline{72.82}   & \textbf{74.45}{\tiny $\pm$0.72}    & \textbf{73.33}{\tiny $\pm$0.58}    & 
\underline{86.52}   & \textbf{56.05}{\tiny $\pm$0.79}    & \textbf{83.34}{\tiny $\pm$0.60} \\
        \bottomrule
    \end{tabular}}
\end{table}

\noindent  \textbf{Comparison methods:} We implemented several leading VLM-based OOD detection methods and a uni-modal method NPOS~\cite{NPOS}, all of which utilized the same pre-trained CLIP as ours for fairness. 
Moreover, we first fine-tune a ViT-B/16 following the setup in previous work~\cite{Arewe}, and then implement several scoring functions (i.e., ViM~\cite{ViM}, GEN~\cite{GEN}, CoPR~\cite{KPCA}) as OOD detectors.

Each model was evaluated with classification accuracy (Acc) on ID testing data, and with the false positive rate of OOD samples when the true positive rate of ID samples is at 95$\%$  (F: FPR95), and the area under the receiver operating characteristic curve (A: AUROC) for OOD detection.

\subsection{Result Analysis}\label{analysis}
\noindent\textbf{Effectiveness evaluation:} 
As shown in Table \ref{tab1}, HVL achieves the best OOD detection performance in AURCO and FPR95 on each dataset, while showing similar average performance on ID classification compared to the best baseline GalLop (86.52$\%$ vs. 86.67$\%$), supporting the superior OOD detection performance from our method.
Especially, HVL achieves good generalization on the pathology benchmark NCT-CRC, indicating that entropy gain remains effective in selecting tissue-related local patch embeddings as pseudo-OOD embeddings.

\begin{figure}[t]
\centering
\includegraphics[width=0.95 \textwidth]{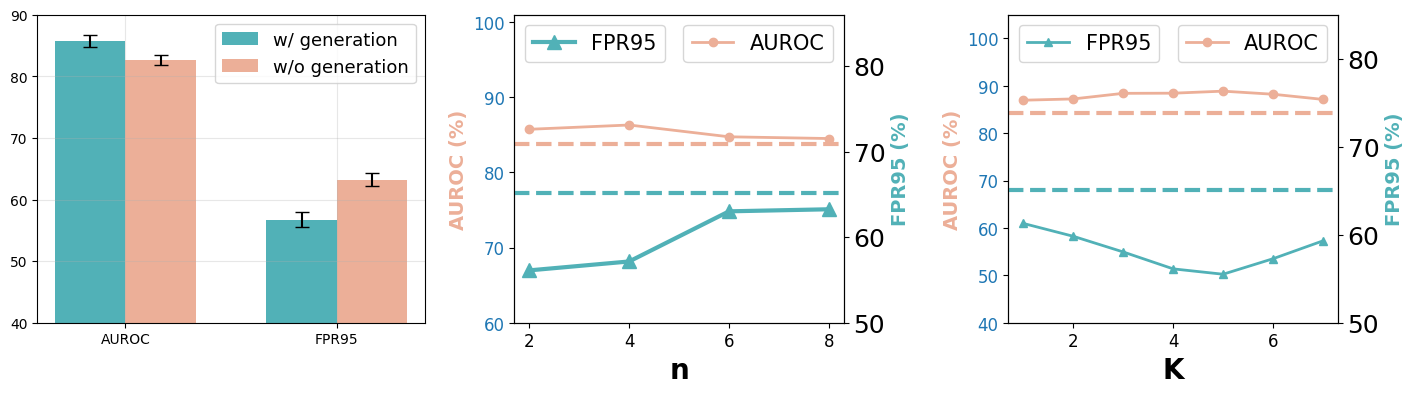}
   \caption{The sensitivity study of hyper-parameters $n$ and $K$ (middle and right), and the ablation study of pseudo-OOD generation at lower scales (left). 
   Dashed lines: performance of the best baseline ViM. Results on ISIC-4 are omitted due to space constraints.}
\label{fig:ablate-sensitive}
\end{figure}

\noindent\textbf{Ablation study:} 
As Table~\ref{table:ablation} (row 2 and row 4) shows, ablation of the loss $\mathcal{L}_{OOD}$  (which includes pseudo-OOD data selected based on $\rm \Delta \rm H (\mathbf{p}^2_j)$) and the cross-scale visual fusion strategy (CSF), respectively, leads to downgraded performance.
Moreover, when replacing the proposed pseudo-OOD selection based on $\rm \Delta \rm H (\mathbf{p}^2_j)$ with the background selection method in previous work~\cite{LoCoOp,SCT},
the performance is also degraded (row 3 vs. row 2), confirming the effectiveness of the proposed selection strategy. Further, when the generation of hard pseudo-OOD embeddings at lower scales is ablated (i.e., $\mathbf q_k^1 = \mathbf q_k^2$ and $\mathbf q_k^0 = \mathbf q_k^2$ in Formula~\ref{eq:refinement}), the OOD detection performance is also significantly decreased (Figure~\ref{fig:ablate-sensitive}, left). 
These results support each key component in our methods helps OOD detection.

\begin{table}[t] 
    \centering
    \caption{Ablation study of HVL on three tasks. `CSF': cross-scale visual fusion. The range of standard deviation is in [0.17, 2.54].}
\label{table:ablation}
    \begin{tabular}{ccccccccccccccc}
        \toprule
\multicolumn{3}{c}{\scalebox{0.9}{\textbf{Components}}} & 
\multicolumn{3}{c}{\scalebox{0.9}{\textbf{Skin40}}}     & 
\multicolumn{3}{c}{\scalebox{0.9}{\textbf{NCT-CRC}}}    & 
\multicolumn{3}{c}{\scalebox{0.9}{\textbf{ISIC-4}}} \\
\cmidrule(lr){1-3} \cmidrule(lr){4-6} \cmidrule(lr){7-9} \cmidrule(lr){10-12}
\textbf{$\mathcal{L}_{OOD}$} &  
\textbf{$\rm \Delta \rm H (\mathbf{p}^2_j)$} & 
CSF& 
Acc & F$\downarrow$ & A$\uparrow$& 
Acc & F$\downarrow$ & A$\uparrow$& 
Acc & F$\downarrow$ & A$\uparrow$\\
\midrule
&       &           &   
65.13   & 86.43     & 70.32 & 
97.12   & 52.63     & 79.86 & 
88.17   & 65.34     & 79.23   \\

 &              &      \checkmark& 
 \textbf{73.32} & 87.50     & 69.71 & 
 98.51          & 49.61     & 87.03 &
 \textbf{88.47} & 64.15     & 81.97\\
 
 \checkmark     &           &       & 
 66.32          & 84.37     & 65.81 & 
 97.72          & 62.42     & 80.39 & 
 88.12          & 63.44     & 82.03\\
 
 \checkmark     & \checkmark &      & 
 67.58          & 80.69     & 69.72 & 
 97.91          & 56.29     & 82.86 & 
 88.36          & 64.89     & 82.47\\
 
\checkmark &   &\checkmark& 
71.94           & 79.49     & 71.35 & 
\textbf{98.59}  & 40.69     & 89.72 & 
88.38           & 61.12     & 84.35\\

\checkmark  &  \checkmark       & \checkmark& 
72.82       & \textbf{74.45}    & \textbf{73.33} & 
98.30       & \textbf{37.59}    & \textbf{90.95} & 
88.43       & \textbf{56.12}    & \textbf{85.75}\\
\bottomrule
    \end{tabular}
\end{table}

\noindent\textbf{Sensitivity study:} 
Figure~\ref{fig:ablate-sensitive} (middle and right) shows our method performs stably well and better than the best baseline ViM when $K$ varies in the large range $[4, 8]$.
Although our method outperforms the best baseline when $n$ varies in $[2,8]$, it can lead to the loss of local information and result in suboptimal performance when $n$ is larger than 4 (i.e., overly small local patches).

\noindent\textbf{Inference cost:}
During inference, original images and patches are encoded in parallel, resulting in latency nearly identical to same-backbone models. HVL incurs only slight memory overhead compared to single-scale methods (2.1GB for LoCoOp vs. 2.8GB for HVL with an image input).

\section{Conclusion}
In this study, we propose a novel VLM-based learning framework for OOD detection. 
The design of cross-scale visual fusion enables the model to capture details of images. 
Moreover, entropy-gain based local region selection and 
hard pseudo-OOD embedding generation improve OOD detection performance across scales.
Extensive experiments validate the effectiveness of our method. 
\begin{sloppypar}
\noindent\textbf{Acknowledgments.} This work is supported in part by the National Natural Science Foundation of China (grant No. 62071502), 
the Major Key Project of PCL (grant No. PCL2023A09), 
and Guangdong Excellent Youth Team Program (grant No. 2023B1515040025).
\end{sloppypar}
\noindent\textbf{Disclosure of Interests.} Authors have no competing interests in the paper.
\bibliographystyle{splncs04}
\bibliography{Paper-0985}
\end{document}